\documentclass[conference]{IEEEtran}
\IEEEoverridecommandlockouts
\usepackage{cite}
\usepackage{amsmath,amssymb,amsfonts}
\usepackage{algorithmic}
\usepackage{graphicx}
\usepackage{textcomp}
\usepackage{xcolor}
\usepackage{bm}
\usepackage{subcaption}
\let\labelindent\relax
\usepackage{enumitem}
\usepackage{cleveref}
\usepackage{float}
\def\BibTeX{{\rm B\kern-.05em{\sc i\kern-.025em b}\kern-.08em
    T\kern-.1667em\lower.7ex\hbox{E}\kern-.125em}}
\usepackage{verbatim}
\usepackage{tikz}
\usetikzlibrary{arrows.meta}
\tikzset{%
  >={Latex[width=2mm,length=2mm]},
            base/.style = {rectangle, rounded corners, draw=black,
                           minimum width=0.5cm, minimum height=0.5cm,
                           text centered},
  activityStarts/.style = {base, fill=blue!30},
       startstops/.style = {base, fill=red!30},
    activityRuns/.style = {base, fill=red!30},
         process/.style = {base, fill=orange!30},
         decision/.style= {diamond, fill=green!30}
}

\newcommand\given[1][]{\;#1\vert\;}

\begin{document}
\usetikzlibrary{shapes.geometric, arrows}
\title{OCTNet: Trajectory Generation in New Environments from Past Experiences
}

\author{Weiming Zhi$^{1,*}$, Tin Lai$^{1,*}$, Lionel Ott$^{1}$, Gilad Francis$^{1}$, Fabio Ramos$^{1,2}$
\thanks{Correspondence to: W. Zhi, {\tt\small weiming.zhi@sydney.edu.au}.}%
\thanks{* Equal Contribution}
\thanks{$^{1}$ School of Computer Science, the University of Sydney, Australia}%
\thanks{$^{2}$ NVIDIA, USA}
}
\maketitle

\begin{abstract}
Being able to safely operate for extended periods of time in dynamic environments is a critical capability for autonomous systems. This generally involves the prediction and understanding of motion patterns of dynamic entities, such as vehicles and people, in the surroundings. Many motion prediction methods in the literature can learn a function, mapping position and time to potential trajectories taken by people or other dynamic entities. However, these predictions depend only on previously observed trajectories, and do not explicitly take into consideration the environment. Trends of motion obtained in one environment are typically specific to that environment, and are not used to better predict motion in other environments. In this paper, we address the problem of generating likely motion dynamics conditioned on the environment, represented as an occupancy map. We introduce the Occupancy Conditional Trajectory Network (OCTNet) framework, capable of generalising the previously observed motion in known environments, to generate trajectories in new environments where no observations of motion has not been observed. OCTNet encodes trajectories as a fixed-sized vector of parameters and utilises neural networks to learn conditional distributions over parameters. We empirically demonstrate our method's ability to generate complex multi-modal trajectory patterns in different environments.

\end{abstract}

\section{Introduction}

Understanding movement trends in dynamic environments is critical for autonomous agents to achieve long-term autonomy. This is further highlighted by the increasing interest in developing autonomous agents capable of coexisting and interacting with humans in a safe and helpful manner, for example, service robots and self-driving vehicles. Humans are adept at anticipating how dynamical objects may move based on the layout of the environment, yet remains non-trivial for robots. Learning to generate likely motion trajectories can allow autonomous agents to anticipate future movement, and better plan in environments with dynamic objects. 

Learning motion trajectories requires the development of predictive models that anticipate complex motions and capture the probabilistic and multi-modal nature of such trajectories. Simple motion prediction approaches, such as constant velocity or acceleration models, involve extrapolating a partially observed trajectory to unseen regions. These methods cannot make use of previous trajectories observed in different environments, and are typically not map-aware \cite{HumanMotionSurvey}. That is, the predicted motion is dependent exclusively on characteristics of the dynamical object, taking neither established paths nor the environment's geometry into account. Advances in machine learning have led to the development of methods which learn the general flow of movement \cite{zhi1,Flow,DirectionalGridMaps}. These methods are map-aware, as they learn the behaviour of motion from all trajectories observed in the environment. Map-aware methods which learn entire trajectories instead of flows have also been developed~\cite{KTM}. However, these methods are restricted to being map-specific, and cannot generalise to environments where trajectories have not been observed.

It is often not possible to observe enough trajectories in a particular environment to build specific flow models of trajectories within a short time. On the other hand, building a static occupancy representation of the environment does not require prior observations of motion trajectories. We are motivated to develop a map-aware motion generation method, capable of generalising to new environments where no motion has been observed, but whose occupancy is known. We contribute a probabilistic generative model, Occupancy Conditional Trajectory Network (OCTNet), capable of generating motion trajectories to new environments by generalising trajectories previously observed in alternative environments.

\begin{figure}[t]
\centering
\begin{subfigure}{.15\textwidth}
  \centering
  \includegraphics[width=\linewidth]{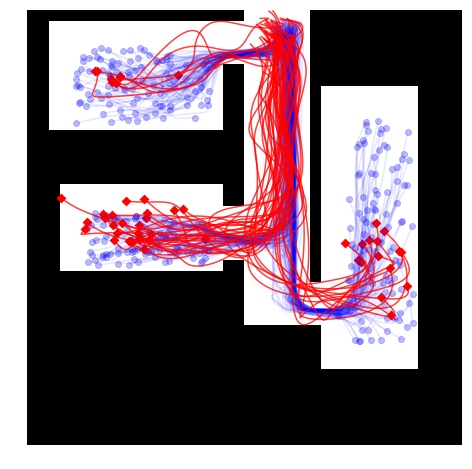}  
\end{subfigure}
\begin{subfigure}{.15\textwidth}
  \centering
  \includegraphics[width=\linewidth]{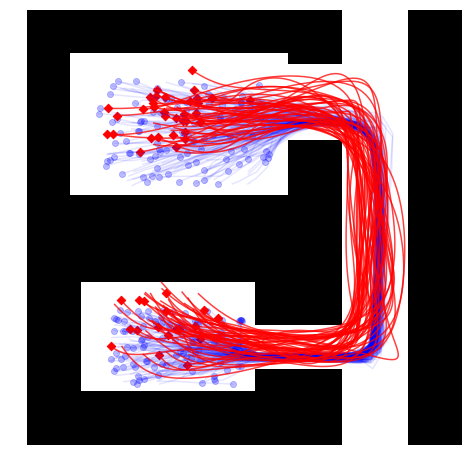}  
\end{subfigure}
\begin{subfigure}{.15\textwidth}
  \centering
  \includegraphics[width=\linewidth]{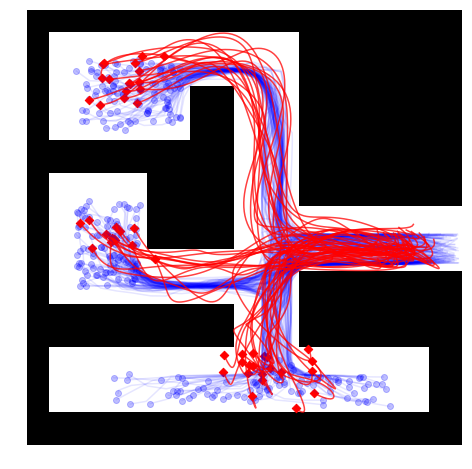}  
\end{subfigure}
\caption{Generated likely motion trajectories (in red) in new environments. End points are indicated by scatter points. OCTNets can capture the probabilistic, multi-modal nature of motion trajectories. Generated trajectories are generalisations of motion trajectories observed in other past environments, and the ground truth trajectories (in blue) hidden during training.}
\label{fig:Pred}
\end{figure}

OCTNet is a generative model with the following desire properties:
\begin{enumerate}[leftmargin=.6cm]
    \item It generalises motion patterns observed in previous environments, to generates motion trajectories in a new environment, where no motion has been observed; 
    \item It effectively models the probabilistic and multi-modal nature of motion generation. Individual trajectories can be generated from the model by sampling from it;
    \item It generates individual trajectories as continuous functions, allowing trajectories to be queried at arbitrary resolution.
\end{enumerate}

\section{Related Work}
OCTNet generates likely motion trajectories in an environment. Generating likely motion trajectories have been studied for a long time. Early simple methods to predict motion are often dynamics-based methods which extrapolate based on the laws of physics \cite{SurveyDynamics}. Examples of dynamics-based methods include constant velocity and constant acceleration models. Dynamics-based methods are utilised in \cite{cycle, ExDynamics2}. The main limitation of these methods is the difficulty of embedding map knowledge, often obtained by observing past trajectories, into the dynamics model. Dynamics models are only able to do very short-term predictions, and do not take into account established paths in the environment. Other attempts at modelling motion trajectories include building dynamic occupancy grid maps based on occupancy data over time \cite{TOG,TempOg,Conditional,GridBased}. However, such methods are typically memory intensive, and can only make short-term predictions.   

Motivated to overcome these limitations, flow-based methods \cite{zhi1,OCallaghan2011LearningNM,embeddings, DirectionalGridMaps,Flow,TAROS} were developed to capture the directional flow in the environment by learning from past observed trajectories, resulting in map-aware \cite{HumanMotionSurvey} models. These methods rely on extracting the movement direction or velocity from past trajectories and then training a mixture of distributions. Motion trajectories can then be generated by starting at an initial location, recursively sampling the distribution of motion directions, and then take a step in the sampled direction. However, forward sampling directional distributions accumulates errors. To address this issue Kernel Trajectory Maps (KTMs) \cite{KTM} were introduced. KTMs modelled entire trajectories as continuous functions, which avoids the shortcomings of forward sampling. However, like earlier flow-based methods \cite{zhi1,OCallaghan2011LearningNM,embeddings, DirectionalGridMaps,Flow,TAROS}, KTMs are limited to extrapolating in environments for which the training trajectories were observed. OCTNet extends ideas from KTMs around generating whole trajectories, and generalises map-aware motion prediction to environments where no trajectories have been observed.

\section{Methodology}

\subsection{Problem Formulation}
This paper addresses the problem of generating likely motion trajectories in a novel environment, based solely on the occupancy representation of the environment.

We assume that we have a dataset containing occupancy representations of an environment and a list of discrete trajectories observed in the environment. We denote the dataset as $\mathcal{D}=\{\mathcal{M}_{n},\{\bm{\xi}_1,\bm{\xi}_2,\ldots\, \bm{\xi}_{P_{n}}\}_{n}\}^{N}_{n=1}$, where $\mathcal{M}_n$ is the occupancy representation, and $\{\bm{\xi}_{1},\bm{\xi}_2,\ldots,\bm{\xi}_{P_{n}}\}$ is the set of $P_n$ trajectories collected in the corresponding environment, where $\bm{\xi}$ contains trajectory waypoint coordinates.

Our proposed method learns a generative model that is capable of sampling from the probability distribution over possible trajectories $\bm{\Xi}^*$, conditioned on an unseen occupancy representation $\mathcal M^*$, i.e.:
\begin{equation} 
    p\left(\bm{\Xi}^{*} \given \mathcal{M}^{*},\mathcal{D}\right),
\end{equation}
where $\bm{\Xi}^{*}$ is the generated trajectory, and $\mathcal{M}^{*}$ is the queried map.

\subsection{Trajectory Representations}
Trajectories in this paper are either \textbf{discrete} or \textbf{continuous}. Recorded trajectory data typically takes the form a sequence of discrete waypoints coordinates, whereas our generated trajectories are continuous functions. The continuous function representation allows for querying at arbitrary resolution without additional interpolation.
\begin{enumerate}[leftmargin=.6cm]
\item \textbf{Discrete trajectories} are represented by an arbitray-length sequence of waypoint coordinates, recorded at fixed time steps. We denote a discrete trajectory, $\bm{\xi}$, with time steps $1 \ldots T$ as, $\bm{\xi}=\{(x_t,y_t)\}_{t=1}^{T}$, where $(x_t,y_t)$ are x,y-coordinates of the dynamic object at time $t$. 
\item \textbf{Continuous trajectories} are smooth continuous functions that map from $[0,1]$ to coordinates. We define a continuous trajectory, $\bm{\Xi}$, as $\bm{\Xi}(\tau)=(x,y)$, where $\tau \in [0,1]$. The time for which the trajectory was recorded is normalised to lie between $0$ and $1$. 
\end{enumerate}
Continuous trajectories can be discretised by querying at uniform intervals between 0 and 1, and retaining the coordinates in sequential order. It is also possible to estimate continuous trajectories of a discrete trajectory. Details of the employed estimation procedure are given in \cref{EmbedTraj}.

\subsection{Overview of OCTNet}
OCTNet generates likely motion trajectories in new environments, generalising trajectories observed in other environments. Generated trajectories are samples from the distribution over possible trajectories, conditional on a given occupancy. We learn a mapping between encodings of occupancy maps to parameters of the required conditional distribution. Realisations of trajectories can then be sampled from the probability distribution.

\begin{figure}[t!]
\centering
\begin{tikzpicture}[node distance=1.5cm,
    scale=0.6, every node/.style={scale=0.6},
    align=center]
  \node (MapIn)[activityStarts] {Input occupancy\\ representations\\ $\{\mathcal{M}_1,\ldots,\mathcal{M}_N\}$};
  \node (GenerateFeatures)     [process, below of=MapIn, yshift=-0.5cm]{Encode each\\ occupancy representation\\ as feature vector\\ of similarities $\bm{\phi}$};
 \node (TrajIn)[activityStarts, right of=MapIn, xshift=2.5cm] {Input discrete\\ trajectories\\ $\{\{\bm{\xi}_{p}\}_{p=1}^{P_1},\ldots, \{\bm{\xi}_{p}\}_{p=1}^{P_{N}}\}$};
\node (GenerateEmbeddings)     [process, below of=TrajIn, yshift=-0.5cm]          {Embed each \\ discrete  trajectory\\ as a fixed-size\\ vector $\bm{w}$ };
\node (TrainModel)      [activityRuns, below of=GenerateFeatures, xshift=2cm, yshift=-0.15cm] {Train MDN to model $p(\bm{w}|\bm{\phi})$};
                                                    
  \draw[->]             (MapIn) -- (GenerateFeatures);
  \draw[->]             (TrajIn) -- (GenerateEmbeddings);
  \draw[->]             (GenerateFeatures) -- (TrainModel);
  \draw[->]             (GenerateEmbeddings) -- (TrainModel);
  \end{tikzpicture} \caption{Process of learning model to generate $p(\bm{w}|\bm{\phi})$}\label{TrainingModel}
\end{figure}
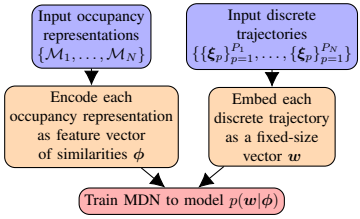

The training process is illustrated in \cref{TrainingModel}, and can be summarised as:
\begin{enumerate}[leftmargin=.6cm]
    \item Construct feature vectors of similarity, $\bm{\phi}$, by calculating the similarity between every relevant occupancy representation. The similarity is found by evaluating Hausdorff distance substitute kernels \cite{DistanceKernel} between occupancy representations. Details in \cref{EncodingOcc}.
    
    \item Obtain a low dimensional embedding $\bm{w}$ of each discrete trajectory. We find the best-fit continuous trajectory to the discrete trajectory, and elements in $\bm{w}$ correspond to weights of predefined radial basis functions of the best-fit trajectory. Details of the embedding in \cref{EmbedTraj}.
    
    \item Learn $p(\bm{w}|\bm{\phi},\mathcal{D})$ using a MDN. Details in \cref{MDN}. 
\end{enumerate}

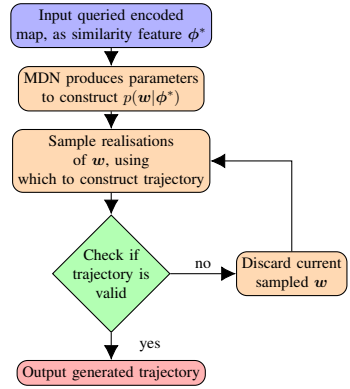
\begin{figure}[t!]
\centering
\begin{tikzpicture}[node distance=1.5cm,
    scale=0.6, every node/.style={scale=0.6},
    align=center]
    \tikzstyle{decision} = [diamond, inner sep=-.3ex, text centered, draw=black, fill=green!30]
    \node (FeatureIn)[activityStarts] {Input queried encoded\\map, as similarity feature $\bm{\phi}^{*}$};
    \node (ObtainProbability)     [process, below of=FeatureIn, yshift=0.1cm]{MDN produces parameters\\ to construct 
    $p(\bm{w}|\bm{\phi}^{*})$};
    \node (SampleW)[process, below of=ObtainProbability, yshift=-0.1cm] {Sample realisations\\ of $\bm{w}$, using\\ which to construct trajectory};
    \node (ObtainTraj)     [decision, below of=SampleW, yshift=-1.0cm]          {Check if \\trajectory is\\ valid};
    \node (Discard)[process, right of=ObtainTraj, xshift=2.5cm] {Discard current\\ sampled $\bm{w}$};
    \node (Finish)     [activityRuns, below of=ObtainTraj, yshift=-0.7cm]          {Output generated trajectory};   
  \draw[->]             (FeatureIn) -- (ObtainProbability);
  \draw[->]             (ObtainProbability) -- (SampleW);
  \draw[->]             (SampleW) -- (ObtainTraj);
  \draw[->]             (ObtainTraj)-- node[anchor=south]{no}(Discard);
  \draw[->]             (Discard) |- (SampleW);
 \draw[->]             (ObtainTraj) -- node[anchor=west, xshift=.5cm]{yes}(Finish);
  \end{tikzpicture} \caption{Process of generating trajectories}\label{QueryModel}
\end{figure}

A brief overview of the generative process is illustrated in \cref{QueryModel}. After the model has been trained, we can input a feature vector $\bm{\phi}^{*}$, associated with the occupancy map of a new environment, to obtain $p(\bm{w}|\bm{\phi}^{*})$. Vectors of $\bm{w}$ can be sampled from $p(\bm{w}|\bm{\phi}^{*})$, and each sample $\bm{w}$ can be used to obtain a continuous trajectory $\bm{\Xi}$. We obtain a distribution over trajectories, and continuous trajectories can be generated by sampling realisations of trajectories from the distribution. As there are no explicit constraints to prevent trajectories from overlapping with occupied regions, we check and only accept valid trajectories to output.

\subsection{Non-parametric Encoding of Environmental Occupancy}\label{EncodingOcc}
We encode the occupancy representation of a given environment as a vector of similarities between the environment, and all environments. 
We define the similarity function, in a similar fashion to the dissimilarity measures described in \cite{DissimilarityKernel}. We substitute the \emph{Hausdorff distance} into a \emph{distance substitute kernel} \cite{DistanceKernel}, to arrive at our similarity function. The Hausdorff distance measures the distance between two finite sets of points, and is commonly used to compare images \cite{HDistImage}.

Given two sets of points $A=\{a_1, a_2, \ldots, a_n\}$ and $B=\{b_1, b_2,\ldots, b_m\}$, and in general $n$ and $m$ are not required to be equal, the one-sided Hausdorff distance between the two sets is defined as:
\begin{equation}
    \hat{\delta}_{H}(A,B)=\max_{a\in A}\min_{b \in B}||a-b||.
\end{equation}
The one-sided Hausdorff distance is not symmetric, we enforce symmetry by taking the average of $\hat{\delta}_{H}(A,B)$ and $\hat{\delta}_{H}(B,A)$, i.e.:
\begin{equation}
    \delta_{H}(A,B)=\frac{1}{2}(\hat{\delta}_{H}(A,B)+\hat{\delta}_{H}(B,A)).
\end{equation}
We can then define a similarity function between two sets $A$ and $B$, analogous to a distance substitute kernel described in \cite{DistanceKernel}, as:
\begin{equation}
    S_{H}(A,B)=\exp\Big\{-\frac{\delta_{H}(A,B)^2}{2\ell_{H}}\Big\},
\end{equation}
where $\ell_{H}$ is a length scale hyper-parameter. We assume that occupancy representations in the dataset are binarised occupancy grid maps. The $n^{th}$ map from the dataset can be represented as a set of occupied locations, $\mathcal{M}_n=\{(x_i,y_i)\}^{I_{n}}_{i=1}$, where there are $I_n$ occupied coordinates given. A Gram matrix of the similarity function evaluated between each map is obtained. The $n^{th}$ row of matrix is a feature vector, $\bm{\phi}_n$, for the map, $\mathcal{M}_n$, we can write this as:
\begin{equation}
    \begin{bmatrix}
    \bm{\phi_1}\\
    \vdots\\
    \bm{\phi_N}
    \end{bmatrix}
    =
    \begin{bmatrix} 
    S_{H}(\bm{\mathcal{M}}_1,\bm{\mathcal{M}_1}) & \ldots & S_{H}(\bm{\mathcal{M}}_1,\bm{\mathcal{M}}_{N}) \\
    \vdots & \ddots & \vdots\\
    S_{H}(\bm{\mathcal{M}}_N,\bm{\mathcal{M}}_1), & \ldots, & S_{H}(\bm{\mathcal{M}}_N,\bm{\mathcal{M}}_{N}).
    \end{bmatrix}. 
\end{equation}
This is equivalent to constructing a kernel Gram matrix between each occupancy representation. However, in this work we treat each row of the matrix as a feature vector. For every map $\mathcal{M}$ in our dataset, there is a corresponding vector of similarities $\bm{\phi}\in\mathbb{R}^{N}$. The process of encoding occupancy information is non-parametric, as for each new data point considered, the length of $\bm{\phi}$ will grow. However, as it is difficult to obtain a real-world dataset with a large number of occupancy maps, with associated motion trajectories in the environment, the number of occupancy maps is typically not large. An alternative parametric model would be to consider comparing only against a subset of occupancy representations, rather than comparing against all other occupancy representations. This is  similar to the concept of representative trajectories in \cite{KTM}. The parametric formulation may increase scalability by sacrificing performance.

\subsection{Embedding Discrete Trajectories}\label{EmbedTraj}
We embed the discrete trajectories as a fixed length vector, by considering the best-fit continuous trajectory. The elements of the vector correspond to weights of fixed basis functions which reconstructs a continuous trajectory that best fits the discrete trajectory. The process of finding the best-fit continuous trajectory of a discrete trajectory is explained below.

We define a normalised timestep parameter $\tau \in [0,1]$. Trajectories in the dataset can have arbitrary timesteps, and $\tau$ indicates the proportion of timestep. $\tau=0$ and $\tau=1$ refer to the first and last timesteps respectively in the discrete trajectory. A continuous trajectory can be modelled by functions, $x(\tau)$ and $y(\tau)$, which map from $\tau$ to the x and y coordinates of the trajectory. We model $x(\tau)$ and $y(\tau)$ by weighted sums of fixed radial basis functions centred on evenly spaced $\tau$ values. Suppose we have a discrete trajectory $\bm{\xi}=\{(x_t,y_t)\}_{t=1}^{T}$, the weights that best fit a given discrete trajectory can be found by solving a pair of Kernel Ridge Regression (KRR) problems, defined as:
\begin{align}\label{KRR}
    \arg&\min_{\bm{w}_x}\sum^{T}_{t=1}(x_t-\bm{w}_{x}^{T}\bm{k}(\tau_t))^2+\lambda||\bm{w}_x||^{2},\\
    \arg&\min_{\bm{w}_y}\sum^{T}_{t=1}(y_t-\bm{w}_{y}^{T}\bm{k}(\tau_t))^2+\lambda||\bm{w}_y||^{2},
\end{align}
where $\tau_t=\frac{t}{T}$ is the normalised time parameter, $\lambda$ is the ridge regularisation parameter, $\bm{k}(\tau_t)$ contains the radial basis function values evaluated at $\tau_t$, obtained by: 
\begin{equation}
    \bm{k}(\tau)=k(\tau,\bm{\hat{\tau}})=[k(\tau,\hat{\tau}_1),k(\tau,\hat{\tau}_2),\ldots,k(\tau,\hat{\tau}_M)]^{T},
\end{equation}
where $\bm{\hat{\tau}}=[\hat{\tau}_1,\hat{\tau}_2,\ldots,\hat{\tau}_M]$ is a vector of $\tau$ values to centre the stationary radial basis functions. In this work, we investigate using the squared exponential basis function, as it is smooth and the default in many kernel based methods. Hence, our basis function is defined by
\begin{align}\label{BasisF}
    \bm{k}(\tau)&=k(\tau,\bm{\hat{\tau}})\\
    &=\Big[-\frac{||\tau-\hat{\tau}_1||}{2\ell_{b}},-\frac{||\tau-\hat{\tau}_2||}{2\ell_{b}},\ldots,-\frac{||\tau-\hat{\tau}_M||}{2\ell_{b}}\Big]^{T},
\end{align}
where $\ell_b$ is the length scale hyper-parameter of the square exponential functions. 

After evaluating \cref{BasisF} to obtain $\bm{k}(\tau_t)$ for each $\tau_t$ considered, we can solve the KRR \cref{KRR}, by computing:
\begin{align}
    \bm{w}_x&=\Big(\lambda\bm{I}+\sum_{t=1}^{T}\bm{k}(\tau_t)^{T}\bm{k}(\tau_t)\Big)^{-1}\Big(\sum_{t=1}^{T}x_{t}\bm{k}(\tau_t)\Big),\\
    \bm{w}_y&=\Big(\lambda\bm{I}+\sum_{t=1}^{T}\bm{k}(\tau_t)^{T}\bm{k}(\tau_t)\Big)^{-1}\Big(\sum_{t=1}^{T}y_{t}\bm{k}(\tau_t)\Big),
\end{align}
where $\bm{I}$ is an identity matrix. We denote the concatenation of $\bm{w}_x$ and $\bm{w}_y$ as $\bm{w}\in\mathbb{R}^{2M}$, where $M$ is the number of basis functions. Every discrete trajectory $\bm{\xi}=\{(x_t,y_t)\}_{t=1}^{T}$ can be converted a corresponding $\bm{w}\in\mathbb{R}^{2M}$, and typically $2M<<T$. We can recover functions that map from an arbitrary queried normalised time parameter $\tau^*\in[0,1]$ to the trajectory coordinates, by $x(\tau^*)=\bm{w}_x^{T}\bm{k}(\tau^*)$ and $y(\tau^*)=\bm{w}_y^{T}\bm{k}(\tau^*)$.

\subsection{Learning a Mixture of Stochastic Processes}\label{MDN}
Each occupancy representation is encoded as a vector of similarities $\bm{\phi}$, and the trajectories observed in the map is embedded as a collection of fixed length weight vectors, $\{\bm{w}_1,\bm{w}_2,\dots,\bm{w}_p\}$, where each $\bm w$ represents a trajectory. We aim to train a neural network to generate $p(\bm{w}|\bm{\phi})$.

Mixture density networks (MDN) \cite{Bishop1994MixtureDN} are a class of neural networks capable of representing conditional distributions. We slightly modify the classical MDN described in \cite{Bishop1994MixtureDN} to learn a mixture of vectors of conditional distributions, corresponding to the conditional distribution for each element in $\bm{w}$. 

To capture the multi-modality of trajectory distributions, we model the conditional distribution $p(\bm{w}|\bm{\phi})$ as a mixture of $Q$ vectors of distributions, which we call components. This can be written as,
\begin{equation}
    p(\bm{w}|\bm{\phi})=\sum_{q=1}^{Q}\alpha_{q}p_{q}(\bm{w}|\bm{\phi}),
\end{equation}
where $p_{q}(\bm{w}|\bm{\phi})$ denotes the $q^{th}$ component, and $\alpha_q$ is the associated component weight. We model each element in the vector with a symmetric distribution. We investigate modelling the distribution over each element in the vector with:
\begin{enumerate}[leftmargin=.48cm]
    \item \textbf{Normal distribution}: The normal distribution is the least-informative default distribution, and is also used in the original formulation of MDNs \cite{Bishop1994MixtureDN}. Under this assumption, we can write each component of the conditional distribution as:
    \begin{equation}\label{NormalAss}
        p_{q}(\bm{w}|\bm{\phi})=\prod_{m=1}^{2M}\frac{1}{\sqrt{2\pi\sigma_{q,m}^2}}\exp\Big\{-\frac{(w_{m}-\mu_{q,m})^{2}}{2\sigma_{q,m}^{2}}\Big\},
    \end{equation}
    where a single component of the mixture $p_{q}(\bm{w}|\bm{\phi})$ can be parameterised by a vector of means, $\bm{\mu}_q=[\mu_{q,1},\mu_{q,2},\ldots,\mu_{q,2M}]$, and a vector of standard deviations, $\bm{\sigma}_q=[\sigma_{q,1},\sigma_{q,2},\ldots,\sigma_{q,2M}]$.
    
    \item \textbf{Laplace distribution}: For many problems a mixture of Laplace distributions was found to provide marginally better performance than a mixture of normal distributions \cite{MDNABrandoMasterThesis}. Under this assumption, we can write each component of the conditional distribution as:
    \begin{equation}\label{LaplaceAss}
        p_{q}(\bm{w}|\bm{\phi})=\prod_{m=1}^{2M}\frac{1}{2b_{q,m}}\exp\Big\{-\frac{|w_{m}-\mu_{q,m}|}{b_{q,m}}\Big\},
    \end{equation} 
    where the component can be parameterised by a vector of means, $\bm{\mu}_q=[\mu_{q,1},\mu_{q,2},\ldots,\mu_{q,2M}]$, and a vector of scale parameters, $\bm{b}_q=[b_{q,1},b_{q,2},\ldots,b_{q,2M}]$. The variance and scale parameter are related by $\sigma^2=2b^2$.  
\end{enumerate}
We can then write the negative log-likelihood loss function over $N$ maps, and $P_n$ trajectories observed in the environment corresponding to the $n^{th}$ map in the dataset as:
\begin{equation}\label{lossF}
    \mathcal{L}=-\log\Big[\prod^{N}_{n=1}\prod^{P_n}_{p=1}\sum^{Q}_{q=1}\alpha_{q}p_{q}(\bm{w}|\bm{\phi})\Big],
\end{equation}
where $p_{q}(\bm{w}|\bm{\phi})$ are given by one of \crefrange{NormalAss}{LaplaceAss}, depending on the assumption of distribution over each element in vector $\bm{w}$. The component weight $\alpha_q$ is associated with the $q^{th}$ component in the mixture. The standard MDN constraints are applied using the activation functions highlighted in \cite{Bishop1994MixtureDN}. This includes:
\begin{enumerate}[leftmargin=.6cm]
\item $\sum^{Q}_{q=1}\alpha_q=1$, such that component weights sum up to one, by applying the softmax activation function on associated network outputs;
\item $\sigma_{q,m}, b_{q,m}\geq0$, standard deviation or scale parameters are non-negative, by applying an exponential activation function on associated network outputs.
\end{enumerate}
\begin{table}[]
\centering
\begin{tabular}{|c|c|c|}
\hline
\multicolumn{3}{|c|}{Input $\bm{\phi}$}                                                                                                                                                                                 \\ \hline
\multicolumn{3}{|c|}{Dense (500 units), ReLU activation}                                                                                                                                                    \\ \hline
\multicolumn{3}{|c|}{Batch Normalisation}                                                                                                                                                                   \\ \hline
\multicolumn{3}{|c|}{Dense (500 units), ReLU activation}                                                                                                                                                    \\ \hline
\multicolumn{3}{|c|}{Dropout (0.25 rate)}                                                                                                                                                                   \\ \hline
\multicolumn{3}{|c|}{Dense (500 units), ReLU activation}                                                                                                                                                    \\ \hline
\multicolumn{3}{|c|}{Dropout (0.25 rate)}                                                                                                                                                                   \\ \hline
\multicolumn{3}{|c|}{Dense (500 units), ReLU activation}                                                                                                                                                    \\ \hline
\multicolumn{3}{|c|}{Dropout (0.25 rate)}                                                                                                                                                                   \\ \hline
\multicolumn{3}{|c|}{Dense (500 units), ReLU activation}                                                                                                                                                    \\ \hline
Dense ($2M\times Q$ units) & \begin{tabular}[c]{@{}c@{}}Dense ($2M\times Q$ units), \\ Exponential activation\end{tabular} & \begin{tabular}[c]{@{}c@{}}Dense ($Q$ units), \\ Softmax activation\end{tabular} \\ \hline
Output $\mu_{q,m}$        & Output $\sigma_{q,m}$ or $b_{q,m}$                                                           & Output $\alpha_{q}$                                                              \\ \hline
\end{tabular}
\caption {The neural network architecture to learn $\alpha_{q}$, $\mu_{q,m}$, and $\sigma_{q,m}$ or $b_{q,m}$ from the feature vector $\bm{\phi}$}\label{architecture}
\end{table}
Using the neural network with architecture illustrated in \cref{architecture} and minimising the loss function defined in \cref{lossF}, we can learn a model that maps from the feature vector of similarities $\bm{\phi}$ to the parameters required to construct $p(\bm{w}|\bm{\phi})$. As a distribution is estimated for each of the elements in weight vector, $\bm{w}$, the predicted $p(\bm{w}|\bm{\phi})$ results in a mixture of discrete processes, where each realisation is a vector of $\bm{w}$. We can also obtain $p(\bm{\Xi}(\tau)|\bm{\phi})=p(\bm{w}^{T}\bm{k}(\tau)|\bm{\phi})$, which is represented by a mixture of continuous stochastic process, with each realisation being a continuous trajectory. $\bm{k}$ denotes the basis functions outlined in \cref{EmbedTraj}.

\subsection{Trajectory Generation}
\begin{figure}[h]
    \centering
    \begin{subfigure}{.38\linewidth}
        \centering
        \vspace{1em}
        \includegraphics[width=\linewidth]{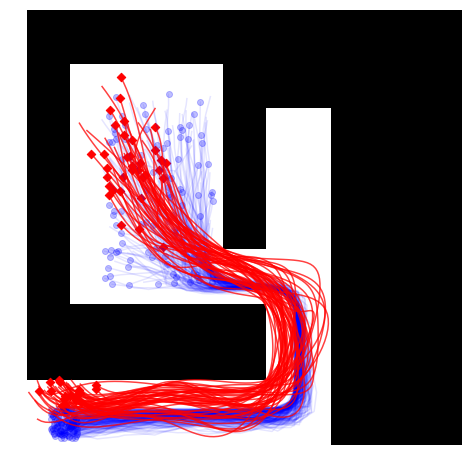}  
    \end{subfigure}
    \hspace{1em}
    \begin{subfigure}[t]{.42\linewidth}
        \centering
        \includegraphics[trim={0 1.5em 0 0},clip,width=\linewidth]{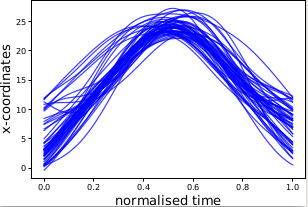} 
        \includegraphics[width=\linewidth]{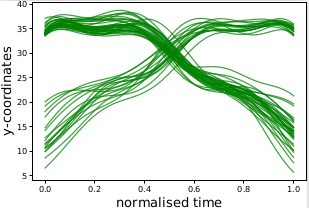}  
    \end{subfigure}
    \caption{(Right) 50 sampled valid $x(\tau)$ and $y(\tau)$  (top, bottom respectively); (Left) corresponding trajectories overlaid on the map, with markers indicating the endpoints. The hidden ground truth trajectories are under-laid in blue. We see that OCTNet is capable of generating different groups of trajectories, and trajectories that start from inside the room and end in the corridor, as well as those moving in the opposite direction.} \label{functionswithT}
\end{figure}

We can generate trajectories in environments with no observed trajectories, by generalising past experiences of observed trajectories in alternative environments. We start by inputting $\bm{\phi}^*$, the feature vector of similarities of the map we wish to query, $\mathcal{M}^{*}$, into the neural network detailed in \cref{MDN}. We obtain parameters, that define distributions over each element in $\bm{w}$. Realisations of $\bm{w}$ can be sampled randomly from the predicted $p(\bm{w}|\bm{\phi}^{*})$, and a possible continuous trajectory, $\bm{\Xi}$, can be found by evaluating $\bm{\Xi}(\tau)=[x(\tau),y(\tau)]=[\bm{w}_x\bm{k}(\tau), \bm{w}_y\bm{k}(\tau)]$, where $\bm{k}(\tau)$ gives the basis function evaluations given in \cref{EmbedTraj}. As there are no explicit constraints in the MDN to prevent the generation of trajectories which overlap with occupied regions, we apply rejection sampling. We query evenly spaced $\tau$ values and check whether $[x(\tau),y(\tau)]$ is occupied against the map $\mathcal{M}^{*}$. If a point on the possible trajectory is occupied, we reject the trajectory, and re-sample $\bm{w}$ from $p(\bm{w}|\bm{\phi}^{*})$. Otherwise, we accept the possible trajectory. \Cref{functionswithT} shows 50 generated trajectories along with plots of $x(\tau)$ and $y(\tau)$. We clearly see that the trajectories generated can belong to different groupings -- one group of trajectories starts from inside the room and exit into the corridor, while the other starts in the corridor and end in the room. The hidden ground truth trajectories are under-laid in blue.

\section{Experiments and Results}
\subsection{Dataset and Metrics}
\begin{figure}[h]
\centering
\begin{subfigure}{.12\textwidth}
  \centering
  \includegraphics[width=\linewidth]{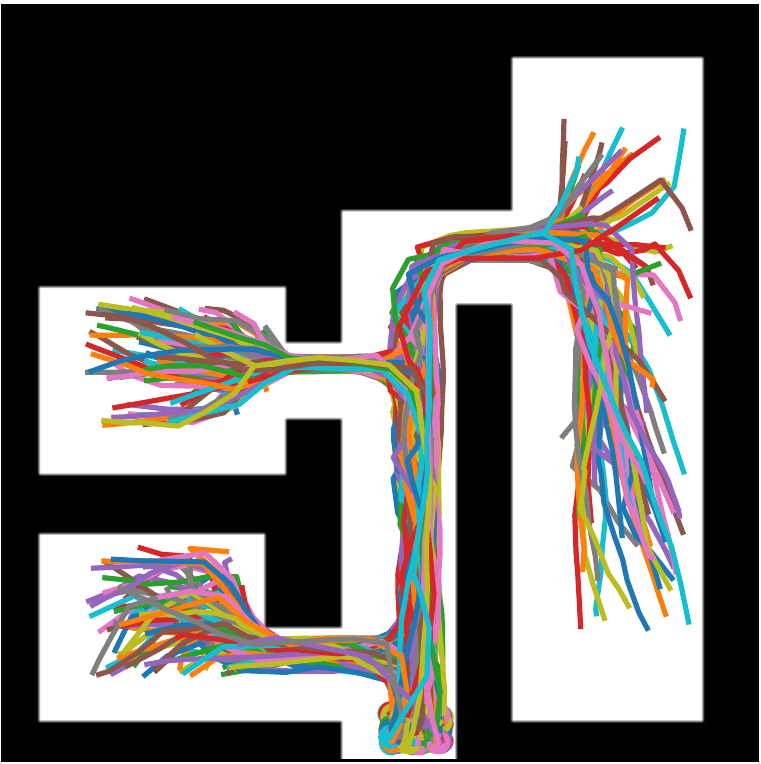}  
\end{subfigure}
\begin{subfigure}{.12\textwidth}
  \centering
  \includegraphics[width=\linewidth]{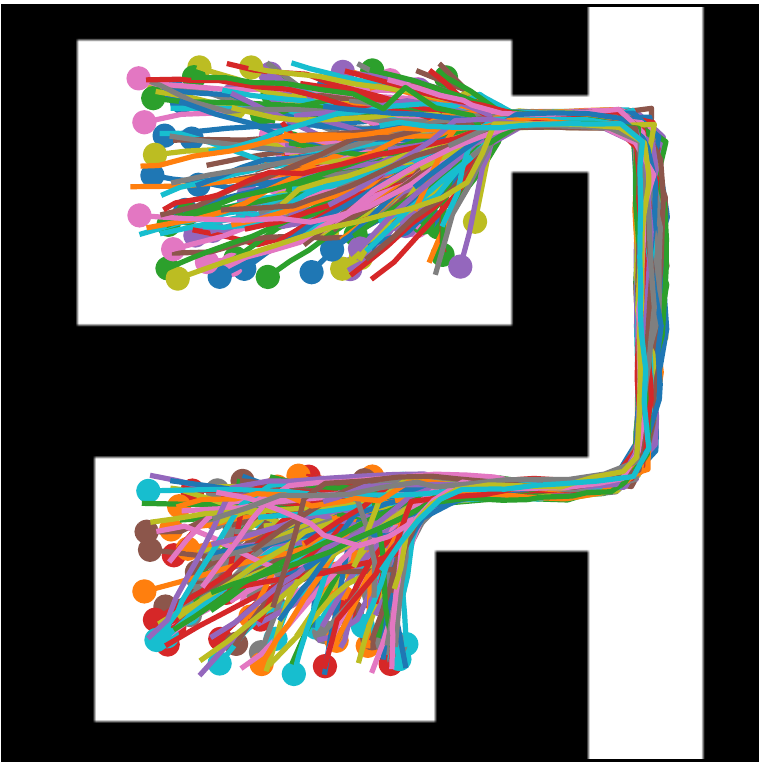}  
\end{subfigure}
\begin{subfigure}{.12\textwidth}
  \centering
  \includegraphics[width=\linewidth]{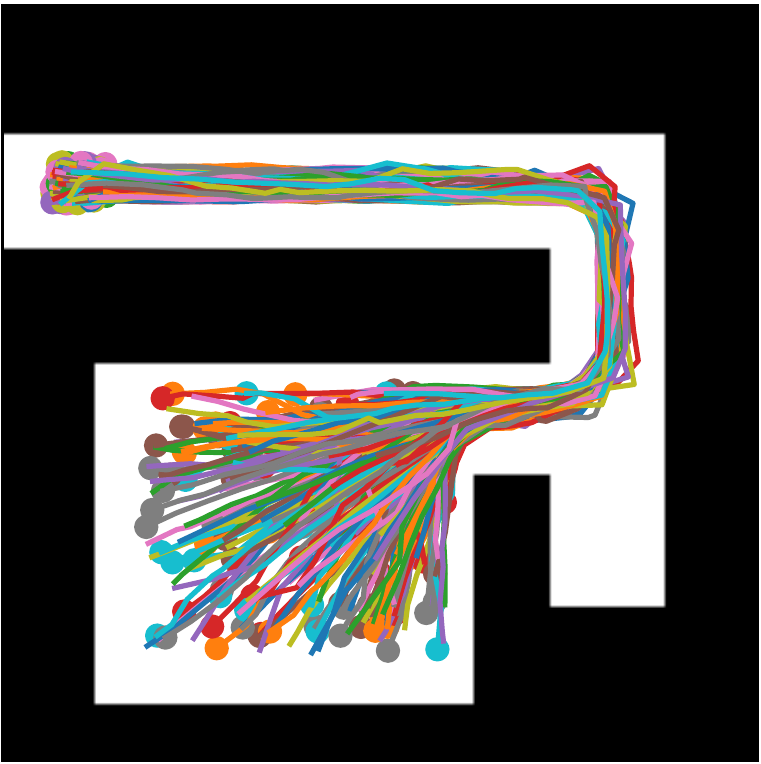}  
\end{subfigure}
\caption{Occupancy maps in the Occ-Traj 120 dataset, along with associated trajectories.}\label{datasetfig}
\end{figure}

Training an OCTNet requires a dataset containing occupancy maps of multiple environments along with observed trajectories in each environment. To the best of our knowledge, there exists no real-world dataset of sufficient size with occupancy data and trajectories observed in different environments. Therefore, we conduct our experiments with the simulated \textit{Occ-Traj 120} dataset \cite{occtraj}. This dataset contains 120 binary occupancy grid maps of indoor environments with rooms and corridors, as well as simulated motion trajectories. Examples of maps and trajectories are illustrated in \cref{datasetfig}.

We evaluate the generated trajectories against a test set with hidden ground truth trajectories. Continuous trajectories outputted are discretised for evaluation by querying at uniform intervals. Due to the probabilistic and multi-modal nature of our output, the metric used is \emph{minimum trajectory distance} (MTD), and is defined by:
\begin{equation}
    MTD = \min_{i=1,\ldots, P}{\mathcal{D}(\xi_{gen},\xi_{i})},
\end{equation}
where $P$ denotes the number of trajectories observed in the environment, with $i$ indexing each trajectory, and $\mathcal{D}(\xi_{gen},\xi_{i})$ is a distance measure of trajectory distance between the generated $\xi_{gen}$ and a ground truth trajectory $\xi_i$. In our evaluations the \textbf{Hausdoff distance}, the \textbf{discrete Frech\'et distance}, and the \textbf{Dynamic Time Wrapping (DTW) Euclidean distance} are considered. These trajectory distances are commonly used in distance-based trajectory clustering to quantify the dissimilarity between trajectories, and a review of these distances can be found in \cite{traj_dist}. Intuitively, the MTD measures the distance between the generated $\xi_{gen}$, and the most similar ground truth trajectory.    

\subsection{Choice of Distribution in Mixture Model}
During the training of OCTNet, distributions over every element of vector $\bm{w}$ are approximated. In each component of the mixture, a class of distributions is taken to be the prior probability distribution. We investigate the performance of assuming normal and Laplace distributions as priors.

OCTNets with normal and Laplace distributions over elements in $\bm{w}$ are trained on 80\% of the maps with associated trajectories in the Occ-Traj 120 dataset, with the remaining 20\% of maps as test. We select the length scale hyper-parameters $\ell_H=50$ and $\ell_b=5$, and train the networks for 10 epochs. 

Performance is shown in \cref{NormalvLaplace}. As demonstrated by evaluating MTD with all three trajectory distances, assuming Laplace distributions as priors over elements in $\bm{w}$ results in generated trajectories which are relatively more similar to the ground truth trajectories. 

The choice of distribution is connected to how trajectories are distributed in the environment. Our results demonstrate that the Laplace distribution assumption leads to stronger results for the Occ-Traj 120 dataset, which contains indoor occupancy maps. However, the most suitable prior may be different in other classes of environment, such as outdoor environments. The OCTNet framework does not limit the choice of prior distributions. Other distributions with closed form probability density functions can be chosen as the prior probability distribution over elements. Our results show the choice of distribution for priors can affect the quality of trajectories, so cross validation methods could be used to guide the choice of prior distributions.

\subsection{Quality of Trajectories Compared to Other Generative Models}
To the best of our knowledge, there exists no other generative methods specifically developed to generate trajectories conditional on occupancy maps. Hence, we evaluate the performance of our generative model, OCTNet, against other popular generative models with slight modifications to generate trajectories. 

We evaluate the performance of Generative adversarial networks (GAN) \cite{Gans} and Conditional Variational Autoencoders (CVAE) \cite{CVAE}, trained to generate vector of weights $\bm{w}$. This is the same vector of weights, $\bm{w}$, OCTNets use to represent trajectories. The GAN and CVAE models are trained for 300 epochs.
\begin{enumerate}[leftmargin=.48cm]
    \item GAN: GANs are generative models that have achieved success in many generative tasks \cite{Gans2, GANimageGen}. We train GANs to generate $\bm{w}$. Trajectories are obtained by evaluating $\bm{\Xi}(\tau)=[x(\tau),y(\tau)]=[\bm{w}_x\bm{k}(\tau), \bm{w}_y\bm{k}(\tau)]$, similar to OCTNets. As we are conditioning on grid maps and outputting weights, the discriminator uses convolutional layers to encode information for prediction; whereas the generator samples a latent variable $\mathbf{\hat{z}}$ with dimensions of 100, concatenated with the map for conditioning. It uses five dense layers to output vector $\bm{w}$,;
    \item CVAE: Similar to the GAN model and OCTNet, we train a CVAE~\cite{CVAE} to generate predictions of $\bm{w}$. The hyperparameters, such as the dimension of $\mathbf{\hat{z}}$, are chosen to be the same as the GAN model. CVAE differs from a GAN, as it utilises the reparametrisation trick to generate structured output prediction through Gaussian latent variables.

\begin{figure}[h]
\centering
\begin{subfigure}[t]{.25\linewidth}
    \includegraphics[trim={0 -1.5em 0 0},clip,width=\linewidth]{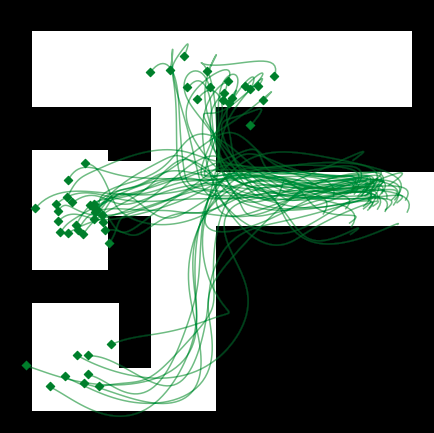}
    \includegraphics[width=\linewidth]{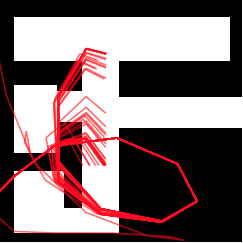}  
\end{subfigure}
\begin{subfigure}[t]{.25\linewidth}
    \includegraphics[trim={0 -1.5em 0 0},clip,width=\linewidth]{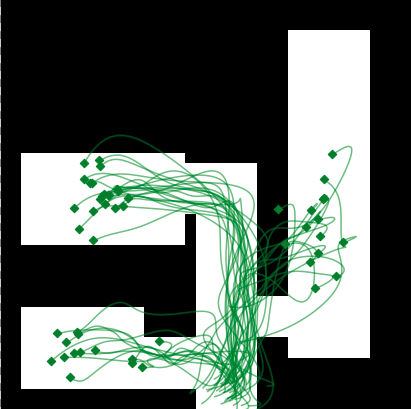}
    \includegraphics[width=\linewidth]{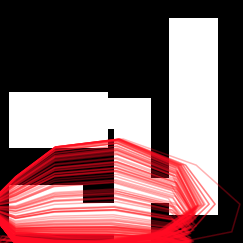}  
\end{subfigure}
\begin{subfigure}[t]{.25\linewidth}
    \includegraphics[trim={0 -1.5em 0 0},clip,width=\linewidth]{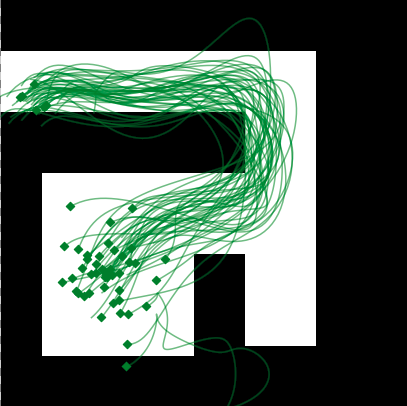}
    \includegraphics[width=\linewidth]{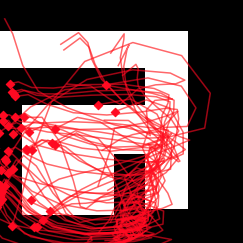}
\end{subfigure}

\caption{Examples of 50 trajectory generated from trained models, conditioned on unseen test maps, without discarding any invalid ones. (Top row, in green) {\color{green!60!black}OCTNet}, (Bottom row, in red) {\color{red}GAN} model.}\label{ExampleGANS}\vspace*{-.3cm}
\end{figure}
\end{enumerate}
In both GAN and CVAE models, we input binary occupancy maps as images during training. It is often not possible to generate a valid trajectory with GAN or CVAE models used for comparisons in reasonable time, as these trajectories would overlap with occupied regions of the map. In these cases, we generate 3000 trajectories for each map using the GAN or CVAE, and select the trajectory with the minimum overlap with occupied regions, as a proportion of the entire trajectory. Comparatively, OCTNet roughly accepts and outputs one out of every three sampled trajectories as valid. Trajectories generated by the OCTNET and GAN, without rejection sampling, are shown in \cref{ExampleGANS}. We see that even without rejection sampling, trajectories generated by OCTNet follow the structure of the environment closely.

\begin{table}[h]
\centering
\begin{tabular}{|l|lll|}
\hline
                     & Hausdoff & Frechet & DTW    \\ \hline
Normal Distribution  & 1.98     & 2.13    & 97.28  \\
Laplace Distribution & $\bm{1.86}$     & $\bm{2.00}$    & $\bm{93.10}$  \\
GANs        & 11.79   & 16.66   & 1147.40\\ 
CVAE        & 9.48 & 14.67 & 751.46\\\hline
\end{tabular}
\caption{Performances of different variants of OCTNet and other generative methods.}\label{NormalvLaplace}
\end{table}

The performance results of our experiments are tabulated in \cref{NormalvLaplace}, we see that OCTNet variants outperform the other generative models compared, demonstrating the high quality of trajectories generated by OCTNet. In particular, the encoding of each occupancy map as a feature vector, $\bm{\phi}$, capturing similarity between all maps in the dataset, allows for flexible representations even when the number of maps in the dataset is small. The MDN used can capture the multi-modal behaviour of trajectories, while the off-the-shelf generative models struggle.

\section{Conclusion}
We present a novel generative model, OCTNet, capable of producing likely motion trajectories in new environments where no motion has been observed, by generalising from past motion trajectories observed in other environments. The OCTNet encodes maps as a feature vector of similarities, and embeds observed trajectories as fixed-size vectors. A neural network is used to learn conditional distributions over the vectors. Realisations of the vectors can then be sampled from the conditional distribution, and used to reconstruct a generated trajectory from the embedding. We investigate two classes of prior distributions over each element of the embedding vector, and empirically show the strong performance of OCTNet against popular generative methods. Future improvements on OCTNet include incorporating temporal changes in trajectory patterns into the framework. Though challenging, there is also a need to collect real-world dataset of occupancy maps with observed trajectories for future research.

\addtolength{\textheight}{-10.1cm}

\bibliographystyle{ieeetr}

\end{document}